# BLOCKCHAIN BASED DECENTRALIZED VOTING SYSTEM SECURITY PERSPECTIVE: SAFE, SECURE FOR DIGITAL VOTING SYSTEM

Jagbeer Singh, Utkarsh Rastogi, Yash Goel, Brijesh Gupta, Utkarsh

Meerut Institute of Engineering and Technology, Meerut.

## Abstract

This research study focuses primarily on Block-Chain-based voting systems, which facilitate participation in and administration of voting for voters, candidates, and officials. Because we used Block-Chain in the backend, which enables everyone to trace vote fraud, our system is incredibly safe. This paper approach any unique identification the Aadhar Card number or an OTP will be generated then user can utilise the voting system to cast his/her vote. A proposal for Bit-coin, a virtual currency system that is decided by a central authority for producing money, transferring ownership, and validating transactions, included the peer-to-peer network in a Block-Chain system, the ledger is duplicated across several, identical databases which is hosted and updated by a different process and all other nodes are updated concurrently if changes made to one node and a transaction occurs, the records of the values and assets are permanently exchanged, Only the user and the system need to be verified no other authentication required. If any transaction carried out on a block chain-based system would be settled in a matter of seconds while still being safe, verifiable, and transparent. Although block-chain technology is the foundation for Bitcoin and other digital currencies but also it may be applied widely to greatly reduce difficulties in many other sectors, Voting is the sector that is battling from a lack of security, centralized-authority, management-issues, and many more despite the fact that transactions are kept in a distributed and safe fashion.

## 1. Introduction to Voting System

Voters' perceptions of elections have been more important in recent years as political parties grow more apart from one another. Every election appears to adopt the slogan of being "the most important in history," and the public reports feeling more upset and concerned about the results. Despite the significance of democratic elections, recent surveys have revealed a decline in public faith in the functioning of the voting system. Only 20% of the population, down 17% from the same poll taken in January of last year, expresses a high degree of confidence in the U.S. national election system, according to a January ABC/Ipsos poll.

All political parties appear to be affected by the decline in trust. Only 20% of independents and only 30% of Democrats see themselves as "extremely confident" in the nation's electoral processes overall, respectively. Surprisingly, only 13% of Republicans say they are "extremely confidence" in the US electoral system [1][2].

## 1. Issues in Current Voting System

1. In light of the most recent elections in the US, a number of software developers claim that electronic voting machines are vulnerable to malicious programming and that if this happens, any hacker could easily penetrate the system and tamper with the results.
2. Many persons with physical limitations have expressed dissatisfaction with the touch-screen voting system's inability to reliably record their choices. Sometimes it results in the voter accidentally casting their ballot for someone else.
3. People have great confidence in the results despite the drawn-out process of tabulating votes put on paper ballots since even cutting-edge technology can be hacked.
4. The largest advance in technology is the ability of a single virus to wipe out the whole data store, regardless of how much data the device collects. Data loss might occur if the electronic voting devices that were utilised during the elections are damaged.
5. Areas with high humidity and those that often rain are not appropriate for employing electronic voting machines to cast ballots. Electronic voting machines should not be used in these places because of the increased risk of machine damage brought on by the high humidity levels.
6. The majority of the electronic voting devices in use in the nation were built abroad, which means that foreigners have access to the secret codes that operate the machines and can use them to sway the outcome of elections.
7. Although fake display units might be put in electronic voting machines to present altered figures, fraudulent votes could initially be produced from the back end. For this approach to compromise the programme, a hacker is not necessary. These fake display products are easily available on the market.
8. The majority of the country's computerised voting machines lack a means for voters to confirm their identification before casting a ballot, which allows for the possibility of fraudulent voters casting a large number of fake votes.
9. The computerized voting machines also don't provide a slip that shows which candidate was selected once the user presses a button. Vote manipulation in these situations is fairly simple for criminals or hackers. If the machines produced such slips, users might check to see if the number of votes recorded by EVMs matched the information on the slips the voter had received.
10. Electronic voting devices may be tampered with during production, and in such circumstances, manipulating the actual vote does not even require a hacker or virus, Many in the crypto-currency industry have urged for the adoption of Block-Chains to replace the present voting system as the public experiences a crisis of faith in elections.

Many people question if voting will be the next industry that Block-Chain technology transforms after revolutionising sectors like cloud storage, smart contracts, crowd-funding, and healthcare in recent years [3][4].

## 2. research and/or analytical methods

Online voting demands the greatest level of security since it is essential to specific processes and outcomes. Features like tamper-proof functioning, scalability, dependability, and real-time updating should be included in an online voting platform. Trending technology can be used to tackle most security-related problems. Block-Chain uses powerful cryptographic algorithms to guarantee the safe and secure preservation of data. Ethereum gives us access to a virtual computer that creates the conditions for building a Block-Chain space and managing it using smart contracts. For online voting, our suggested solution will integrate all essential and desired features.

In order to solve these issues and guarantee a secure, quick, and accurate voting process, it is crucial to address the voting system's limitations. A platform called Immune Ballot was developed on the Ethereum network to address the serious issues raised above and to update the antiquated voting system.

## 3. findings facts and results

1. Many in the Block-Chain community think that Block-Chain technology can make voting safer, simpler, and more accessible so that more people can fulfil their fundamental democratic responsibility.
2. Since at least 2012, there has been talk about enhancing voting with the technology made popular by bitcoin.
3. Since then, a number of businesses have started creating Block-Chain voting solutions.

### 4. The significance or implications of Voting System

There are a number of crucial processes that many believe would take place in a Block-Chain system, even though developers have envisioned various strategies for applying Block-Chain technology while voting. In a decentralised voting process, the following events would most likely occur:

1. To cast a ballot, a voter must enter their identification.

2. All information is encrypted and kept in transactional form.

3. Each node in the network receives the broadcast of the transaction, which is then promptly confirmed.

4. When the network authorises the transaction, the data is recorded in a block and added to the chain; once added, it cannot be modified or altered.

5. Users may examine the findings and, if they so desire, go back and look at previous transactions [5][6].

## 2. System Model

Voting used to be fairly straightforward hundreds of years ago since governments were not departmentalized and everything was centrally managed by a few notable faces. In the twenty-first century, however, there is a department for every facet of governance. This creates a very complex scenario in which numerous department heads must collaborate to develop a regulatory framework. Following that, a solid but automated system must assure the long-term conformity of voting occurrences.

Block-Chain technology is made up of a single, mutually agreed-upon record of transactions that is shared by millions of nodes. To modify the current data on the network, the hacker or any fraudster would need to reach a consensus, which would imply "forcing" 51% of the total nodes to default at the same time. Because it is practically and computationally (nearly) impossible, the likelihood of a record being modified is close to nil. This is arguably the most well-known feature of this technology that makes it suited for voting systems.

The advantages of the Block-Chain system as a voting system-

1. A Decentralized Database Even if a hacker gets into one node, he won't be able to shut down the entire network. Biometric Verification Block-Chain Technology requires identification verification prior to action 3. Security Cryptographic encryption and the use of private and public vital mechanisms ensure security.

4. Transparency The voting process is transparent while maintaining individual confidentiality.

5. Generous architecture on top of the utilized Block-Chain technology, anyone can develop Use-Cases.

6. Harmless to the ecosystem casting ballot voting forms and coordinated operations cause a ton of emanations. Block-Chain has a significantly lower impact on the environment thanks to new mining techniques [7][8].

b. Prior Research

Decentralization has been around for a long time; nevertheless, Block-Chain innovation is its most effective use. Despite this, despite the fact that Bitcoin was established many years ago, the majority of people assume that Block-Chain technology can only be utilised for financial objectives.

However, when more people began to use it, particularly as Enterprise-Grade Block-Chains such as Ethereum became accessible, everyone saw this technology's great potential and adaptability.

In 2012, a group of Canadian academics attempted to "fortify" the architecture of electronic voting systems by using the intrinsic properties of the Bitcoin network to do "carbon dating" in a digital environment and track the origin of information.

Commit-Coin was the first approach established by the two scientists in this area. They were able to secure everyone's votes and make them irreversible once they were broadcast.

b. The paper's rationale Impact

The US elections in 2016 and 2020 should have demonstrated that there has never been a greater desire in safe and non-manipulable democratic cycles. In this post, we'll show you how to achieve that by using the benefits of Block-Chain technology [9].

The term "democracy" has become a cliché in many countries. On the one hand, governments convince their subjects that they have the ability to "manage" the country, but citizens must wait in line for hours to elect their leaders. Worse, despite all of this effort, the majority of countries remain dubious about vote tampering [10].

It is crucial to recognise that voting methods exist at all levels. Elections occur everywhere, from the selection of a school's head boy or head girl to the election of the president. However, because the national level employs the most modern voting technology (relatively), we can get a clear understanding of the issues that the rest of the country would face if Tier-1 leadership was unable to assure that the people could vote in a transparent way.

Hackers can get access to a voting system before the event begins and continue manipulating the entire process without anybody noticing.

To give you a better picture, imagine a hacker gaining access to a constituency voter database and deleting a few thousand entries. As a result, when these voters walk to the polls, the system will be unable to recognise them. As a result, they will be unable to vote.

Let us also discuss national elections in any country. Thousands of electronic gadgets are involved in every event that employs a range of networks and software. Because of their versatility, they are relatively easy to hack, and once exposed, only the hacker who can remotely manage these devices can do anything with them. It is critical to remember that virtually all traditional voting systems are class-based, which means that if one machine fails, the problem will spread to every device and system on the network.

It means that if a single flaw is exploited, thousands of systems might rapidly begin to fail.

We only discussed challenges with national voting systems above, but small-scale voting systems face the same issues.

d. Description the methodology used

Block-Chain voting-cast is like to traditional analog- voting-cast. The procedures with ideas are the same. A citizen would need to register and demonstrate his/her nationality in a particular authority in order to throw a digital vote. The user's citizenship and identity could then be recorded on the Block-Chain linked with that user's key. A ballot is the next thing a citizen needs to vote for. This most likely would take the form of a unique voting coupon that would be depositedin user-account on the Block-Chain. Additionally, this token would likely have a time limit for voting, after which it either would destroy by itself through a smart contract or cease to be useful.

The ballot, or voting token, must be delivered to a specific address in order to be cast on the Block-Chain. The address that supports which candidate or referendum would be known to voters. A coupon sent to that attends to would be a vote [11].

To facilitate sounds technically straight-forward adequate. The vote is recorded on the Block-Chain, where it is transparent, verifiable, and indestructible. To determine who will win the election, we can easily count the votes. Additionally, we are able to design attractive user interfaces that both conceal and automate the process of sending a token to a particular address. Instead, voters would be presented with a straightforward online interface where they could select a candidate or proposal and click the submit button [12].

e. An outline. Introductions often conclude with an outline

Block-Chain voting is not yet ready for primetime or perfect. However, once it achieves legitimacy, democracy is likely to undergo significant change. An engaged electorate will result from voting procedures that are simpler and more transparent. It could also serve as a reminder of the purpose of representatives—to deliberate on policy full-time and make wise decisions regarding matters that the general public may not be able to fully research.

## 3. Literature Overview-

Before EVMs, ballot papers, which were frequently a little ball or piece of paper used for secret voting, were utilised. The term "polling form" is plagiaristic beginning with Italian word " ballotta " it's refers to little ball which is used in voting and secret ballot cast with voting forms in Venice-Italy. During an election, voters indicate their choices on a printed ballot. The options are then totaled, and the ballot is saved in case the election results are challenged. Voters originally recorded their choices with a tiny ball. Each voter casts only one ballot; none are shared.

A ballot is simply only a piece of paper on which each voter writes the name of the candidate they desire to support in the most basic elections. However, employing pre-printed voting forms during legislative sessions is certainly required to ensure ballot confidentiality. The voter projects their ballot into a case at the voting site. Let us study about the narration of ballotpapers, India's transition to electronic voting machines and Voter verifiable paper audit trail (VVPAT), and the adoption of none of the above (NOTA) in India's election:[13].

PAPER BALLOTS

It is stated that palm leaves were used for town gathering decisions throughout the province of Tamil Nadu during historical times in India, perhaps about 920. Candidates' names were formerly inscribed on palm leaves and placed in a mud pot to be counted after voting. The name of the system was Kudavolai.

Formerly to the advent of electronic voting in the 1990s, India utilised paperballot and human counting but paperballots were extensively condemned due to fraudulent voting and booth capturing, in which party adherents loaded counting boxes with pre-filled phoney votes and captured booths. Because tallying hundreds of millions of individual ballots needed a large amount of post-voting resources, in print paperballots was very costly.

In India, Electronic-Voting using Electronic Voting Machines (EVM) is the most common method for conducting elections. In the 1990s, the state-owned Electronics Corporation of India and Bharat Electronics developed and tested the use of EVMs and electronic voting. They were introduced gradually in Indian elections from 1998 to 2001.

In the 1982 byelection to the North Paravur assembly election-constituency in Kerala, EVMs were utilised for the first time testing purpose of polling stations and later, EVMs were employed in a few Delhi, Madhya Pradesh, and Rajasthan seats as an experiment. In 1999, EVMs was utilised for the initially in the entire state of Goa. In 2003, Electronic Voting Machines (EVMs) was utilised in all by-elections and state elections. The Indian Election-Commission were decided to utilise exclusively and EVMs for Loksabha used in 2004.

An embedded EVM characteristics like as "electronically limiting with speed of casting votes to five/minute," a safety measures "lock-close" mark, an electronic record for "voting signatures and thumb impressions" to verify the voter's uniqueness, and deploying extensive security personnel at each booth have all helped to reduce electoral fraud and abuse, eliminate booth capturing, and create more competitive and equitable elections [14][15].

Indian EVMs are self-contained computers with memory that can only be read and written once. The EVMs are self-contained, battery-powered devices with no networking capabilities. They are made with safe manufacturing practices. They contain no wireless or wired internet interfaces or components.

**Why Current Voting System is awful?**

1. Voters with uniformed biases: Our population votes for the wrong people for the wrong reasons. (The winner is the one who can best orally win votes, not leaders or people with good plans.)

2. In a democracy, ignorance is the most powerful weapon available to the corrupt.) Governmental issues being a Vocation: Instead of trained politicians who want a job, we need passionate leaders who want the best for the country.

3. Sponsorship: The system has been ruined by corporate sponsorship; the more money you accept, the stronger your campaign will be the more favours you will receive if the leader benefits a particular third party and not another association, which is typically the case.

4. Favouritism, by ignoring some candidates and giving others screen time, the media can pick who wins. Furthermore, the editors are free to choose how each candidate is presented. for instance: When Donald Trump performed his most recent political stunt, interviews featured him literally ranting about the birth certificate, and the news was overflowing with stories featuring him discussing the document. There was very little discussion of his actual strategy[16][17].

## 4. Proposed System General Architecture

This Proposed system is the basic approach to deal with the election associated with the Block-Chain Technology. We would be talking about the central level election approach too. This System gives you the basic understanding of approach depending on the size of the election.

A block chain system is combined with a client-server architecture in the proposed system. A smart-phone, computer with a webcam having internet connection are the basic necessities for a voter. For a Large scale elections, we would have a choice of electoral voting machines connected to a Block-Chain Network.

**Parts of the System**

**1 Server-A (Authentication Server) -**

1.1 Basically used for Registration of user.

1.2 Connected with the arbitration server as well as Traditional database.

1.3 On the time of Registration, stores the data of the user in the private cloud database

**Server-B(Arbitration Server)**

4.1 Uses the Authentication Server to validate the user during voting.

4.2 Send the key to the voter's computer to encrypt their vote.

4.3 The arbitration server forwards the vote to the relevant node for inclusion in the Block-Chain network.

## 5. Block-Chain System –

5.1. A form of distributed ledger technology (DLT) based on mathematics, cryptography, and computer science is known as Block-Chain. A Block-Chain is essentially a piece of ever-expanding code that works across a global network of computers known as "nodes."Data or financial transactions are verified, managed, and stored by nodes. Furthermore, financial incentives make it more profitable to support Block-Chain networks than to attempt to manipulate them.

5.2. A Block-Chain network allows users to send transactions from address A to address B. The network's nodes take these transactions and mathematically verify their validity. A "block" is created and the verified transaction is cryptographically linked to the previous one (hashed).A block is added to the Block-Chain when it contains all transactions. Additionally, every transaction becomes immutable, meaning that no one can alter or remove it.

5.3. Block-Chain is a convenient technology that enhances the transparency, immutability, and security of any transactions or procedures. As a result, an electoral Block-Chain voting system is a hot topic of discussion because it has the potential to significantly enhance existing voting infrastructure. However, there are some technical obstacles to overcome before we see a Block-Chain voting system in mainstream electoral polls because Block-Chain is still a relatively niche and developing technology. Universities around the world are publishing a lot of research papers that look into the possibilities of using an electoral Block-Chain voting system in countries. Also, there are only a few active Block-Chain voting system projects that make it possible for various businesses and operations to hold cryptographically secure distributed democratic events.

## 6. Smart Contract-

1. Smart contracts are computer protocols that digitally enable the implementation of an agreement and are kept in public databases.

2.They offer a more secure, quicker, and less expensive way of maintaining and executing contracts.

3.If smart contracts are to be broadly accepted, certain difficulties must be addressed, notably the technical complexity of making updates and the inability to handle complicated transactions.

4. Because smart contracts offer a safe environment, the voting mechanism is less vulnerable to manipulation.

5. Deciphering ledger-protected votes using smart contracts would be incredibly difficult.

Furthermore, smart contracts may increase voter turnover, which has historically been low due to an inefficient system that requires voters to line up, verify their identity, and complete forms. Casting a ballot, when done online with smart contracts, can increase the number of members in a democratic framework.

**5. Database-**

1. In contrast to Block-Chains, the database is a centralized, administrator-controlled ledger.

2. The ability to read and write are just a few of the unique characteristics that databases possess. Write and Read actions can only be performed here by parties with appropriate access.

3. Additionally, databases can store multiple copies of the same data as well as their history. This is accomplished with the assistance of a dependable, centralized authority in charge of the server.

6. **Module Design Phase-1 User Registration Process for Voting-**

Assuming the voting day as 12th of January 2023, So the actual preparation for voting start happening few months ago. A crucial need of elections is voters.

For a transparent election, we need an eligible voters depending upon the nature of elections.

In Registration process, all the users have to create a voting account in order to complete the first step for voting [18][19].

User need to carry identification proofs to verify the user as to fulfil the criteria of the eligibility. The Eligibility criteria is based upon the nature of election.

Details such as Aadhar-Card Number, Name, DOB and a voter’s photo also have been taken from all possible angles along with a 2 minute video clip.

When all these things have been done then verification process occurs where the information has been checked from the government record and when it find the information accurate then the eligibility passed notification is sent to the Aadhar card linked number of the voter.

The TLS v1.2 protocol is used to send all information between the user and the AS to ensure its security.

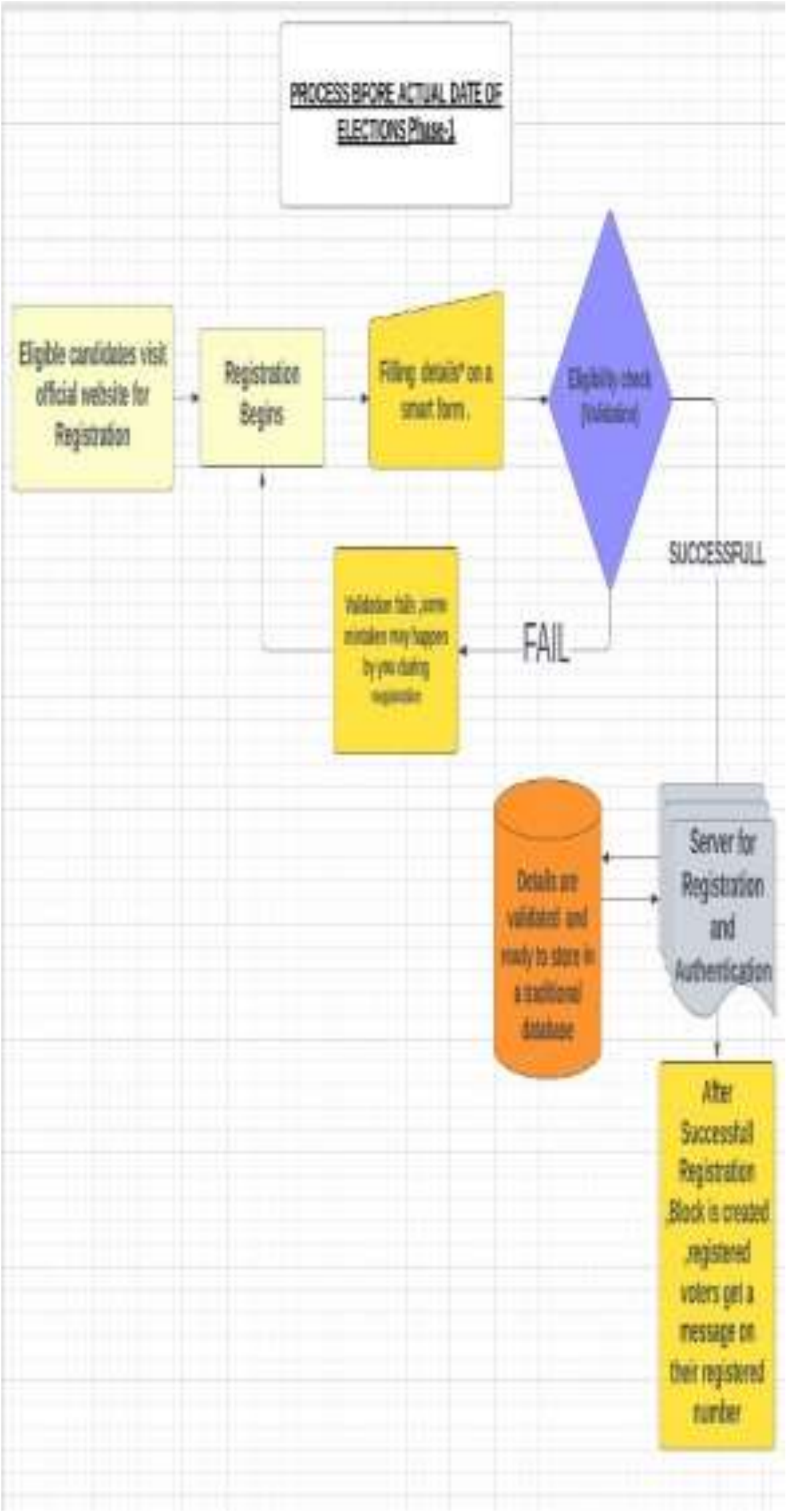

**Phase-2 (On the Election Day):**

1. On the Election day, Eligible voters access the Secure Voting application interface for voting.

2. They enter the Login details to verify themselves.

3. It provides full transparency so that non-eligible voters cannot able to do the vote.

4. Verifying users is the most important task and when they pass that, few images and video has been taken of them to verify and ensure background details etc.

5. Pictures of the user has been continuously until vote has been casted taken of the user, and send to the server to verify the presence of the voter.

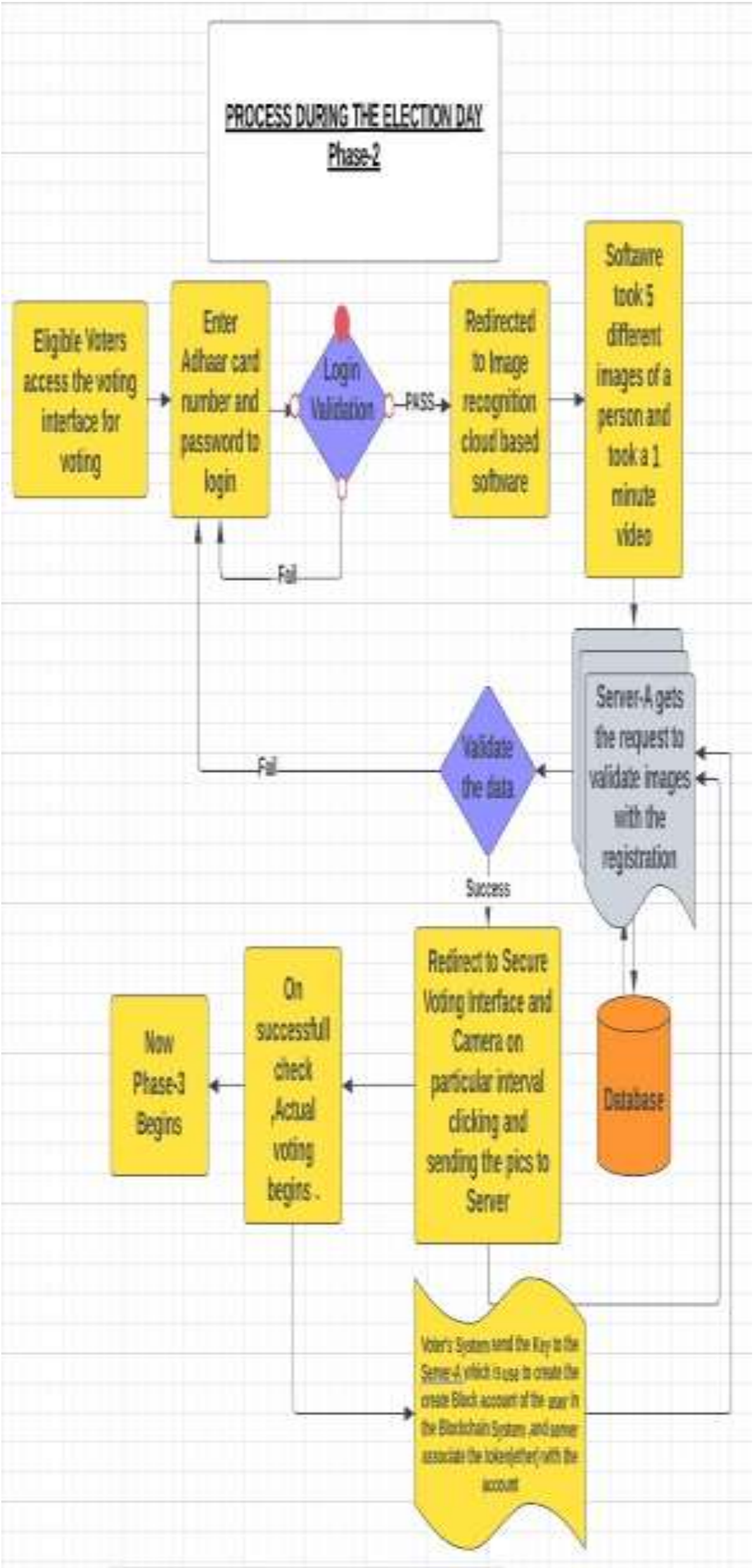

6. On Successful verification of the voter, voter's app sends the key to the server-A where key is associated with the username to create a Block accounts of the user and add it into the Block-Chain System.

7. Server-A, also associate the Ether or token with the user Block account which is used at the time of voting [20].

**Now Phase -3 starts.**

**Phase 3(Voter Casting their Vote)-:**

1. After logging in, the user's system generates a public key and sends it to the Auth Server.

2. The AS would link the key and login.

3. A predetermined amount of ether would be added to the user's account to enable

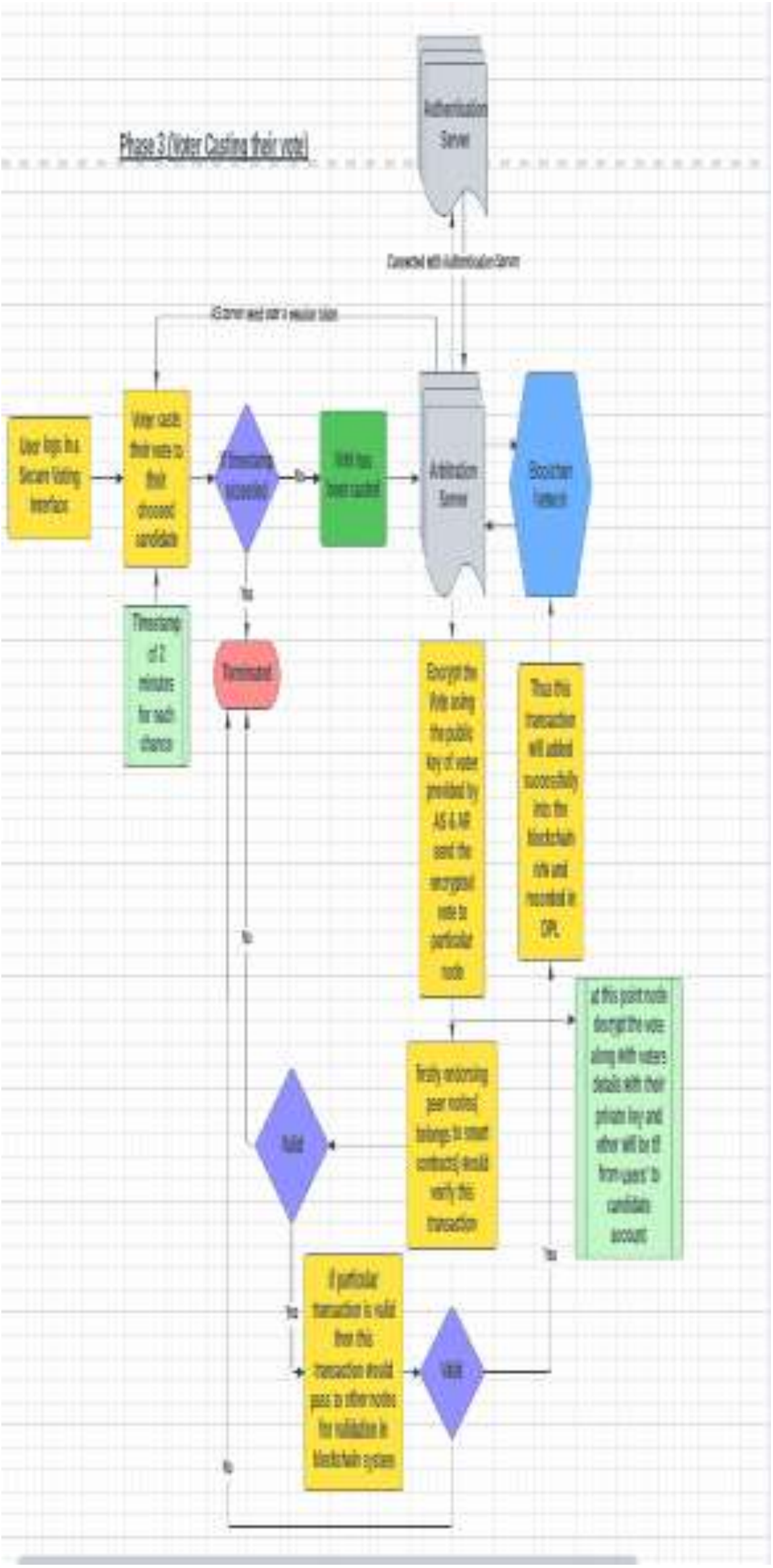

voting after the key was used to create the user's account on the Block-Chain system. The user would be directed to the AR after receiving a session token from the AS.

4. The Modified Needham-Schroeder protocol is used to verify and generate a token following the user's delivery of the session token to the AR and confirmation with AS. These shields against impersonation and man-in-the-middle attacks.

5. Along with the public key of the Block-Chain node to which the user's vote would be sent, the AR would send the user a message to verify their identity. The user would send their encrypted vote using the public key.

6. The vote would be concealed as a result, and the Arbitration server would not be able to see it. The relevant node would get the encrypted vote from the AR.

7. The node then transfers ether from user accounts to candidate accounts after decrypting the vote and obtaining the voter's private key.

8. Each node would confirm the transaction and then validate it using smart contracts.

9. If the transaction is legitimate, it is sent to other Block-Chain system nodes for additional verification.

10. As a result, this transaction will be successfully added to the Block-Chain and logged in DPL.

**Phase 4(After the Election):**

**Verifying the Vote-:**

1. The method used to verify a voter's cast your vote depends on the choice and its category of election. Elections can sometimes produce provisional results, but never be always.

2. The voter must always be informed that their operation former accepted by the Block-Chain structure.

3. One of the Block-Chain's nodes might be made available to the public if an election allows for intermediate results.

4. By inputting their public key on a website resembling "https://Block-Chain.info," users will be able to verify whether their votes were counted.

5. This node was unable to introduce any new transactions into the network. It will be put into action using smart contracts and non-editable transactions. The system's attack surface will be reduced as a result.

6. As a result, the AR is a thin client that uses an intermediate to verify transactions. Here users would be capable to view election results at the conclusion.

Counting the Votes:-

1. Vote counting procedure is rather straightforward.

2. Each voter has a set amount of ether or other money with which to select a candidate.

3. The election is won by the candidate with the most amount of ether on hand.

4. The ether of customers who vowed not to vote would be moved off a Go without Record. This guarantees that their vote won't be compromised.

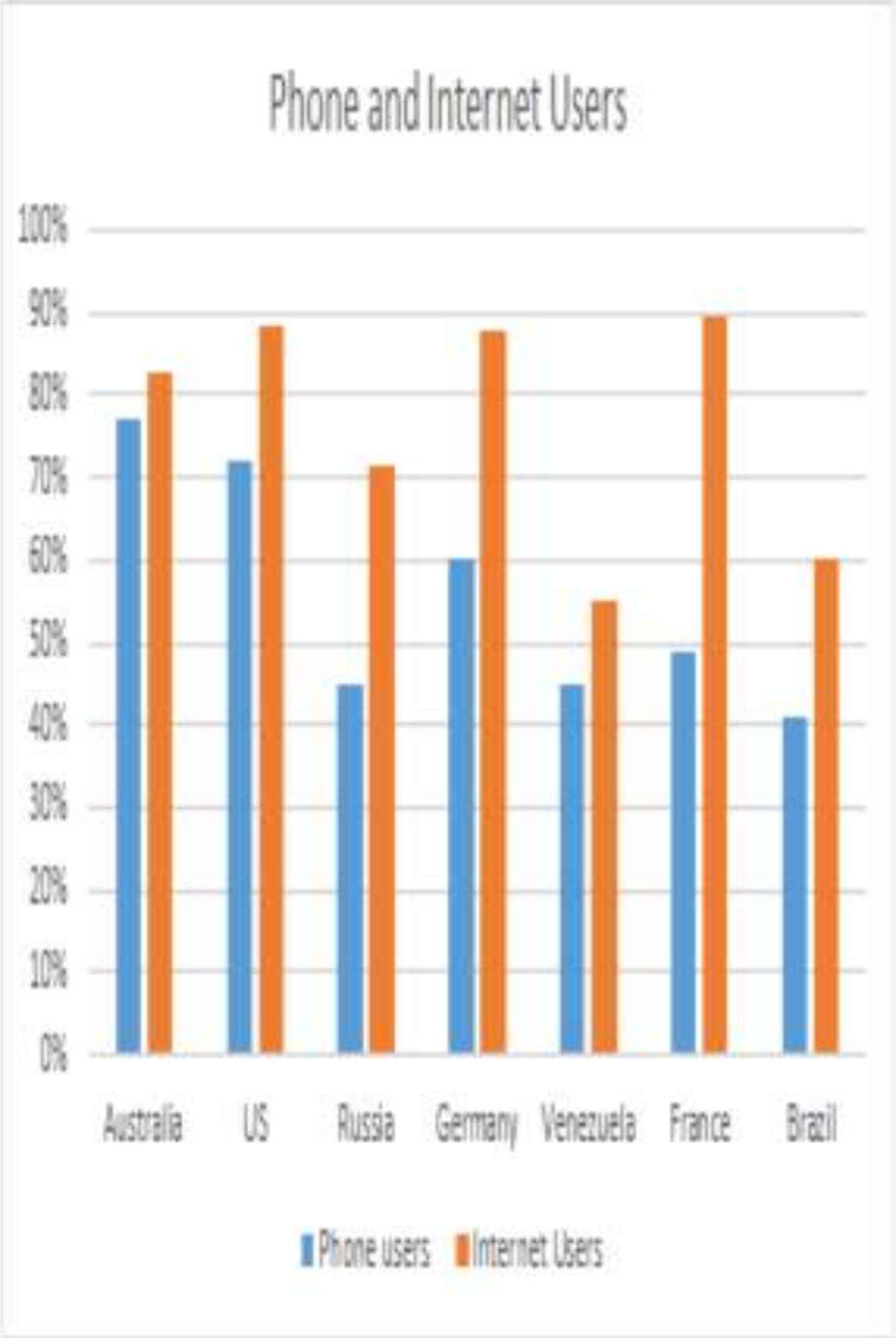

Figure 2: Graph of phone and internet penetration of a country

**Recounting the Votes-:**

1. Election results are occasionally contested. The suggested system makes it easier to solve a variety of problems.

2. Anyone can check to see if their vote was counted by looking at the whole hierarchy related to a particular account source. Users will be able to see the system as a result.

3. Since no one is aware of which user is connected to which account, voting anonymity is protected.

4. Individual Block-Chain transactions' public keys can be used to link accounts in the AS.

5. The list of voters is included in the public key list that was established. The election may be verified by the mapping of this lists to the public-keys connected to every operation.

**Economic advantages of proposed system-**

1. Analyses of the proposed system's costs and benefits. The system would need to be built and tested by 25 people over a 12-month period, which would cost $80 per hour of labour. Customer time cost $4,000,000. Cost of equipment and upkeep for first political race = $100,000,010 for an elector base of per 100 millions (in light of EC2 number cruncher) and (Incorporates server farm costs, network gear, and data transfer capacity)
2. For ensuing political race cycles cost = $50,000,010. A Ballotbased Election would cost $200,000,000 for 100 million voters, or $2 per person. Using the proposed system for the first time, the election would cost $104,000,000,000 to run.

**3. Cost for a country**

The country's internet and smart-phone penetration is depicted in the graph above. For a user to vote, these are the two minimum requirements. Since people can still vote in government buildings like libraries, internet penetration is given more weight. The framework requires the web entrance of a country to be more prominent than or equivalent to 90% to make this venture possible. In order to provide users with an internet connection, the government will need to make excessive investments in infrastructure below that level. Every day, a large number of people worldwide gain access to the internet.

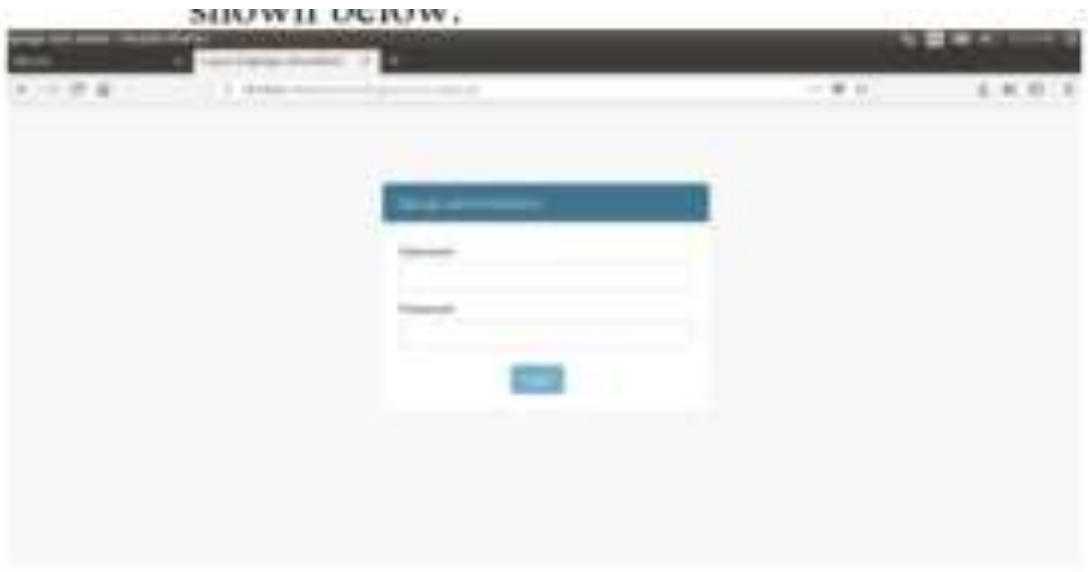

Fig 5 : Admin Panel

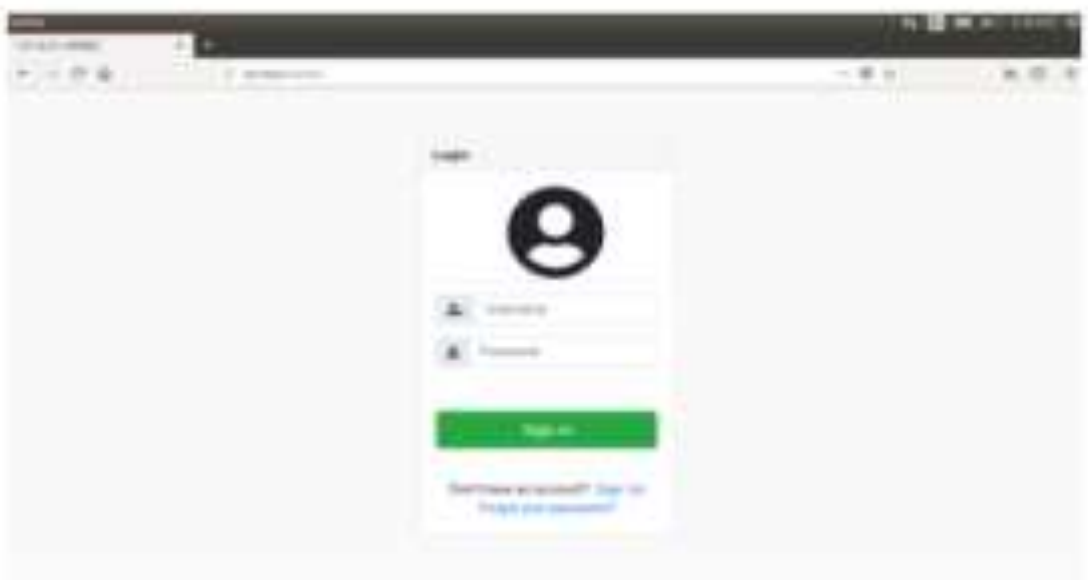

Fig 6 : Voter's Login Form

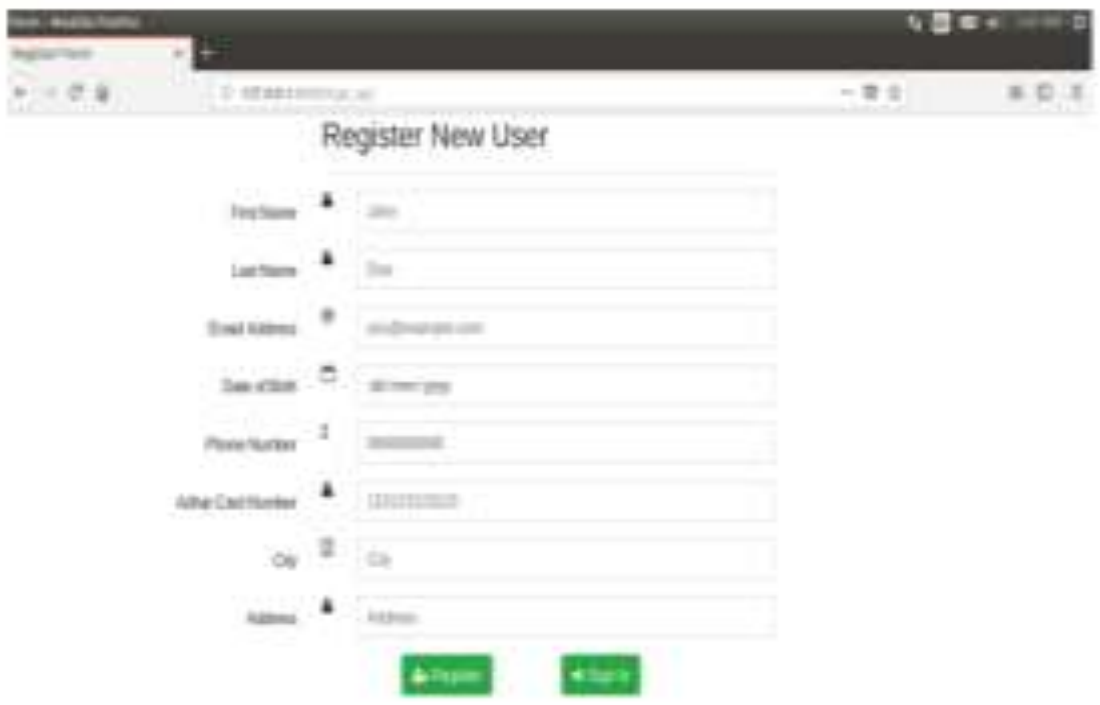

Fig 7 : Voter's Registration Form

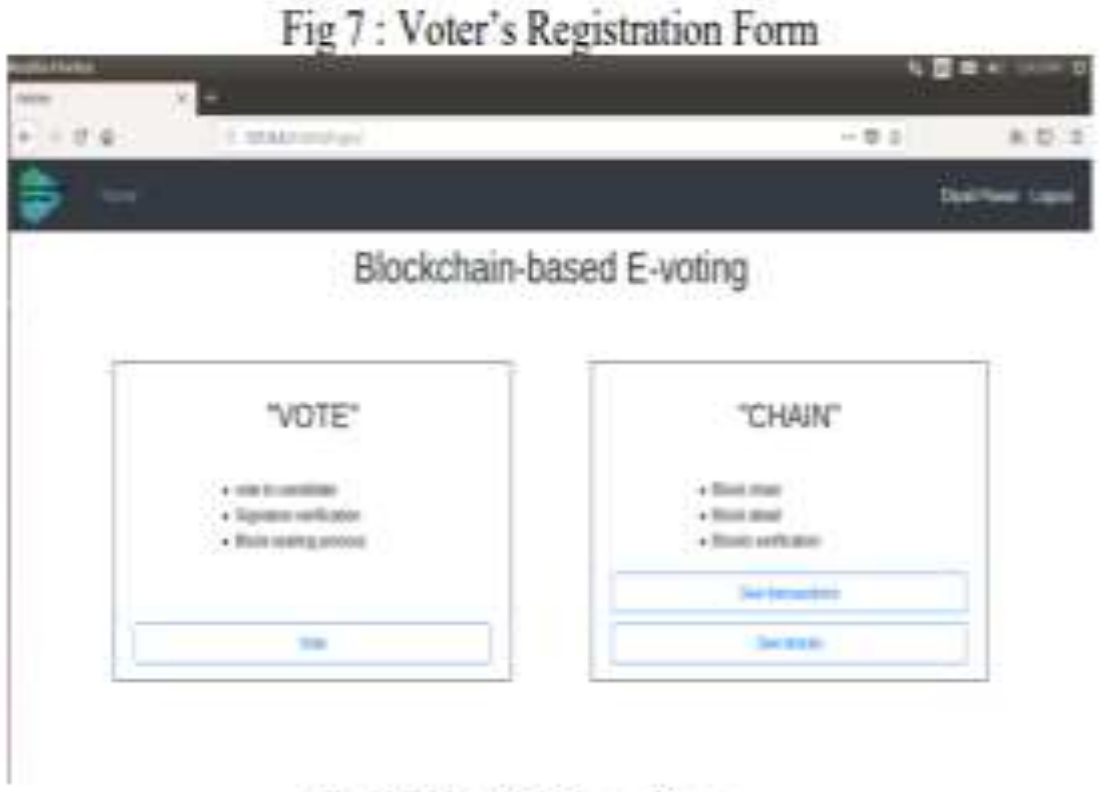

Fig 8: Voter's Home Page

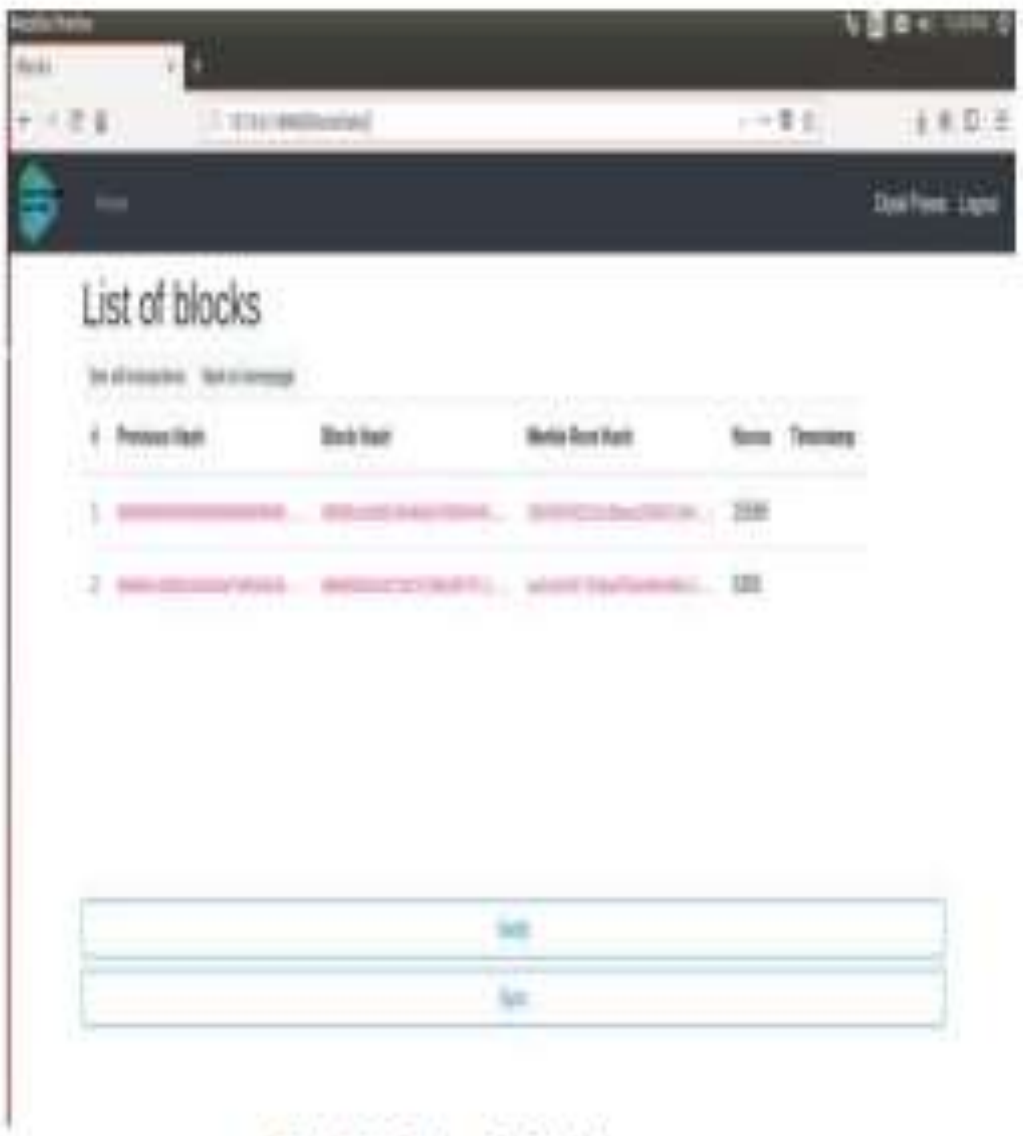

Fig 11 : List Of Blocks

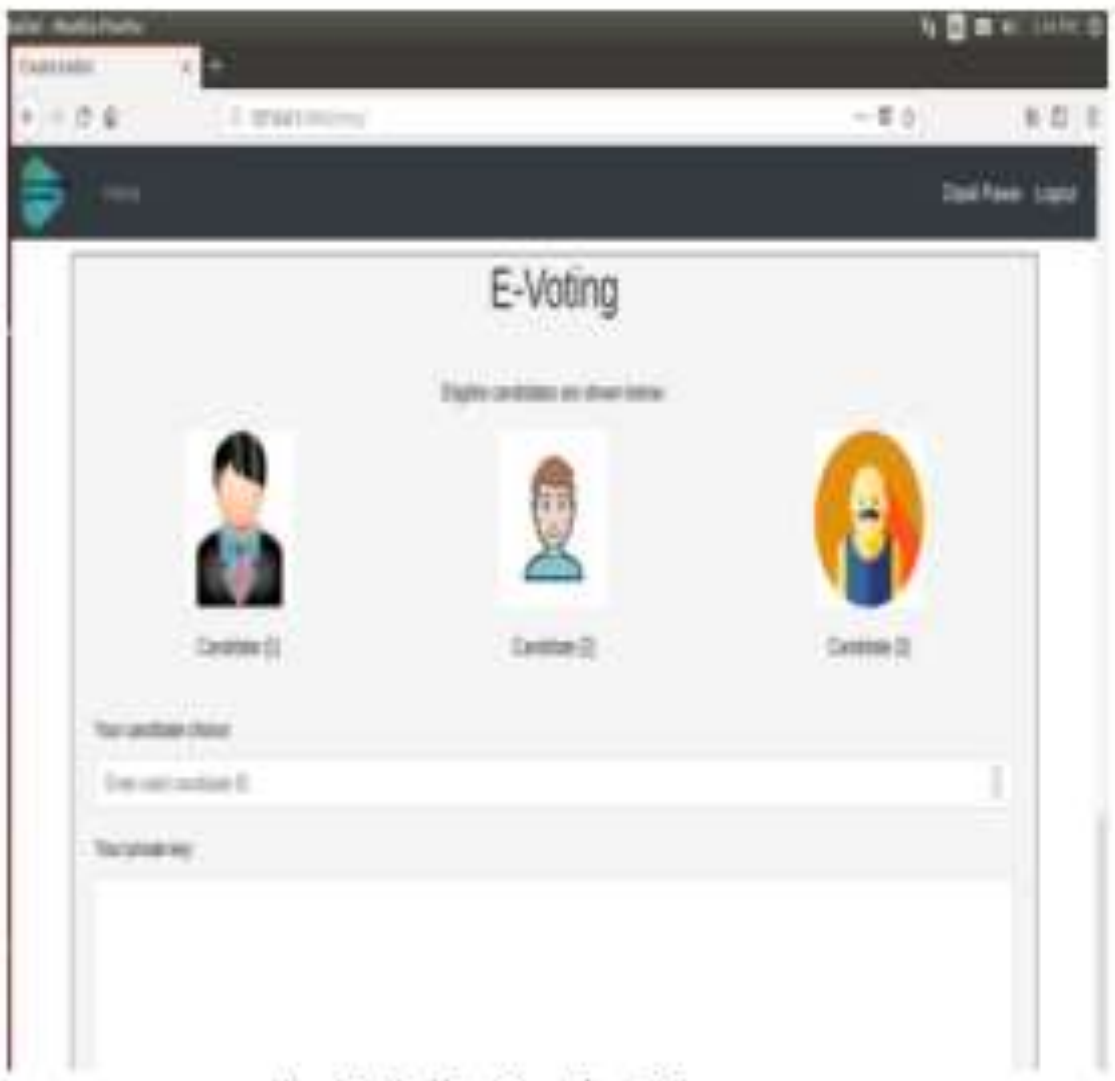

Fig 9 : Ballot Creation Page

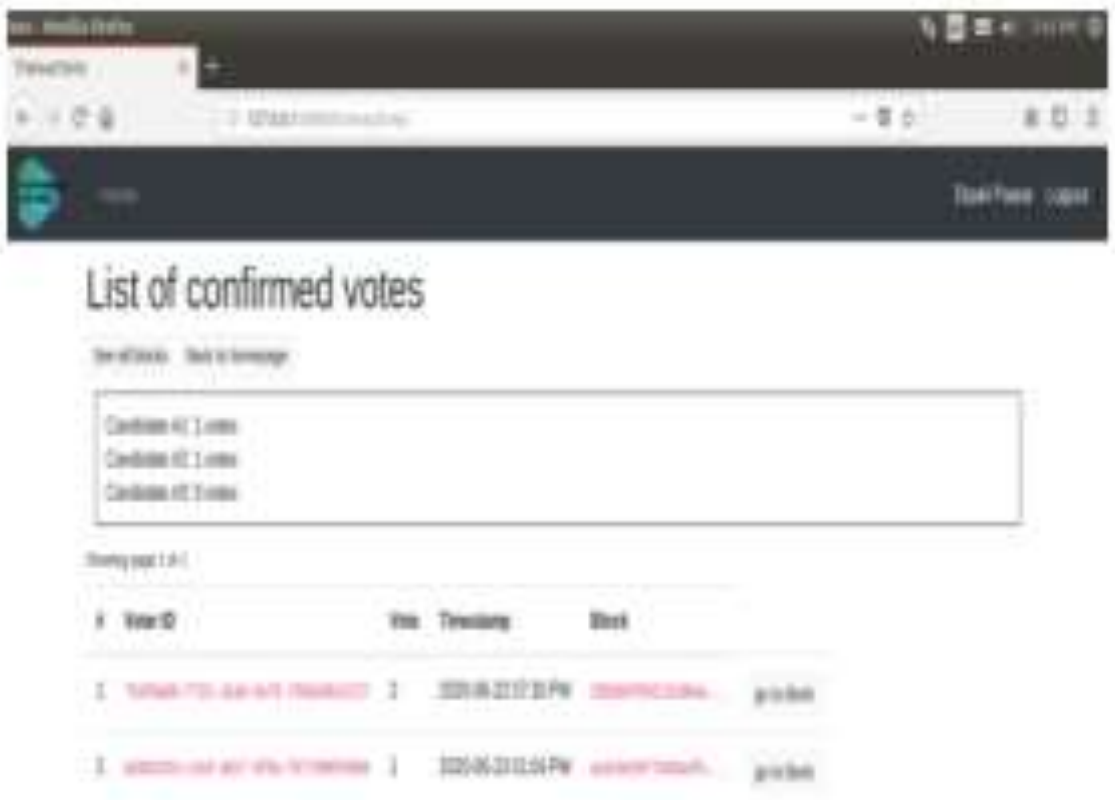

Fig 10 : Transaction Page

## 6. Conclusion

In this paper we introduced a Block-Chain-based electronic voting system which uses smart contracts to ensure voter privacy while simultaneously facilitating cost-effective, secure client elections. Here we demonstrated that Block-Chain technology presents a novel opportunity to avoid the limitations and barriers to the adoption of electronic voting systems, thereby establishing the foundation for the transparency and ensuring election integrity and security. It is possible to send hundreds of transactions onto an Ethereum private Block-Chain each second by utilizing every aspect of the smart contract to lighten the load on the Block-Chain and transparency makes it easier to audit and comprehend elections.

Decentralized networks have these characteristics, which can make elections, especially direct election systems, more democratic. A possible solution would be to base e-voting on Block-Chain technology to make it more open, transparent, and independently auditable. The purpose of this project is to investigate the potential of Block-Chain technology and its application to the e-voting system. The Block-Chain will be distributed in a way that makes it impossible for anyone to corrupt it and will be verifiable by the public.

Voting is an essential component of electing representatives in any democratic nation and nation's policies are formulated by these representatives, who are a part of the government. In various forms, electronic voting has been around for 50 years. With the headway of innovation, the utilization of Block-Chain has expanded in numerous areas going from monetary to non-monetary. When applied to electronic voting, Block-Chain technology has the potential to enhance efficiency while also preserving voters' privacy. It facilitates transactions quickly and securely. Voters are kept out of fraudulent transactions by anonymity. This paper discusses the advantages of incorporating Block-Chain technology into the E-Voting system. However, it might be affected by double spending, which means that the same vote could be cast more than once.